\documentclass[sigconf]{acmart}   



\usepackage{subcaption}

\usepackage{caption}
\usepackage{algorithm}
\usepackage{algorithmic}

\usepackage{arydshln}  

\usepackage{amsfonts}
\usepackage{newfloat}
\usepackage{listings}
\usepackage{pifont}
\usepackage{amssymb}    \usepackage{booktabs}
\newcommand{\cmark}{\checkmark}
\newcommand{\xmark}{\ding{55}}

\AtBeginDocument{%
  }

\copyrightyear{2026}
\acmYear{2026}
\acmDOI{10.1145/3770854.3780189}
\setcopyright{cc}
\setcctype{by}
\acmConference[KDD '26]{Proceedings of the 32nd ACM SIGKDD Conference on Knowledge Discovery and Data Mining V.1}{August 09--13, 2026}{Jeju Island, Republic of Korea}
\acmISBN{979-8-4007-2258-5/2026/08}
\acmBooktitle{Proceedings of the 32nd ACM SIGKDD Conference on Knowledge Discovery and Data Mining V.1 (KDD '26), August 09--13, 2026, Jeju Island, Republic of Korea}
\acmPrice{}

\begin{document}

\title[AnchorGK: Anchor-based Graph Learning Framework for Inductive Spatio-Temporal Kriging]{AnchorGK: Anchor-based Incremental and Stratified Graph Learning Framework for Inductive Spatio-Temporal Kriging}

\author{Xiaobin Ren}
\affiliation{%
\department{School of Computer Science}
\institution{University of Auckland}
  \streetaddress{22 Symonds Street}
  \city{Auckland}
  \country{New Zealand}
  \postcode{1010}
}
\email{xren451@aucklanduni.ac.nz}

\author{Kaiqi Zhao}
\authornote{Kaiqi Zhao is the corresponding author.}
\orcid{0000-0002-0984-1629}
\affiliation{%
\department{Shenzhen Key Laboratory of Internet Information Collaboration}
\institution{Harbin Institute of Technology, Shenzhen}
  \streetaddress{22 Symonds Street}
  \city{Shenzhen}
  \country{China}
  \postcode{518055}
}
\email{zhaokaiqi@hit.edu.cn}

\author{Katerina Ta\v{s}kova}
\orcid{0000-0002-3217-7877}
\affiliation{%
  \department{School of Computer Science}
\institution{University of Auckland}
  \streetaddress{22 Symonds Street}
  \city{Auckland}
  \country{New Zealand}
  \postcode{1010}
}
\email{katerina.taskova@auckland.ac.nz} 

\author{Patricia Riddle}
\orcid{0000-0001-8616-0053}
\affiliation{%
  \department{School of Computer Science}
\institution{University of Auckland}
  \streetaddress{22 Symonds Street}
  \city{Auckland}
  \country{New Zealand}
  \postcode{1010}
}
\email{pat@cs.auckland.ac.nz}



\renewcommand{\shortauthors}{Xiaobin Ren, Kaiqi Zhao, Katerina Taškova, and Patricia Riddle}

\begin{abstract}
 Spatio-temporal kriging is an essential research problem in sensor networks due to the sparsity of deployed sensors.
 While recent studies consider spatial and temporal correlations, they often overlook the sparse spatial distribution of locations and the incomplete features across locations.
To tackle these problems, we propose an 
\underline{Anchor}-based Incremental and Stratified \underline{G}raph Learning Framework for Inductive Spatio-Temporal \underline{K}riging (AnchorGK). AnchorGK introduces anchor locations to enable effective data stratification for accurate kriging. Anchor locations are constructed based on feature availability, and strata are subsequently established based on the anchor locations. This stratification serves two purposes: 1) it ensures that the spatial correlations between unknown areas (no observations) and surrounding known locations are accurately represented and dynamically updated within the graph learning framework, and 2) it facilitates the use of all available features across different strata through a novel incremental representation method. Building on the data stratification, we propose a dual-view graph learning layer that integrates information from relevant features and locations and learns distinct representations for different strata. Finally, kriging is performed based on the obtained strata representations. 
Experimental results on multiple benchmark datasets demonstrate that AnchorGK consistently outperforms existing state-of-the-art methods\footnote{Our codes, datasets, and related materials are given in: https://github.com/xren451/Spatial-interpolation}.
\end{abstract}



\maketitle

\section{Introduction}
Location-based sensing technologies have advanced various research areas, such as geosciences~\cite{GHM}, environmental science~\cite{DAMR}, and Internet of Things~\cite{INCREASE}. High deployment costs lead to low sensor granularity, while varying deployment challenges across sensor types result in inconsistent feature availability~\cite{Imbalancefeatures}. These limitations hinder fine-grained spatio-temporal analyses~\cite{STGNP}.
Spatio-temporal kriging is an essential tool
to increase the spatial resolution by mining data from observed (known) locations to infer information for unobserved (unknown) locations~\cite{KrigingReview}.




\begin{figure}
    \centering
\includegraphics[width=\linewidth]{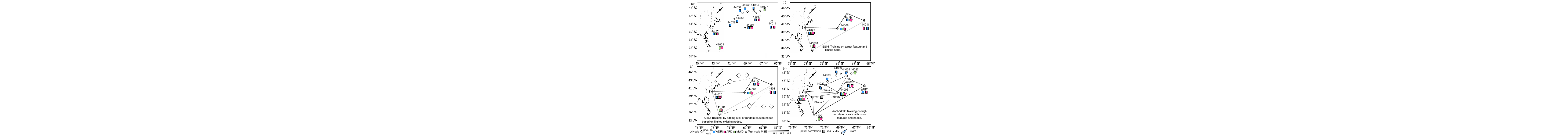}
\caption{Sparse spatial distribution and incomplete observed features on UScoast dataset. (a) Location on the part of UScoast dataset. 
(b) and (c) stands for MSE on MWD (denoted by the colour of nodes) for the decremental training process in SSIN~\cite{SSIN} and the incremental training process in KITS~\cite{KITS}. (d) Incremental training process and MSE on MWD of AnchorGK.}
\label{fig:intro_fig}
\vspace{-2em}
\end{figure}

Traditional spatio-temporal kriging employs statistical methods~\cite{kriging1990,adaptiveIDW}.
Although these methods have proved effective, 
the Gaussian assumption for observations across locations limits their applicability~\cite{KCN, SSIN, KITS}.
Recently, Graph Neural Networks (GNNs) have emerged as a powerful tool to capture complex spatial correlations between locations~\cite{GAT,graphsage}, making them a promising tool for spatio-temporal kriging~\cite{KCN,IGNNK}. However, real-world spatio-temporal data are often \textbf{sparse in space} and have \textbf{incomplete features across locations}. Figure~\ref{fig:intro_fig} (a) illustrates the location of UScoast weather locations and their feature availability. As observed from the figures, certain locations (e.g., locations 41001 and 44025) are sparsely distributed. Besides, some features are available only at some locations, e.g., only wind direction (WDIR) is observed at location 44032, while all features are available at location 44008. Such characteristics of spatio-temporal data present practical challenges for existing approaches~\cite{KCN,SSIN,IGNNK,INCREASE,STGNP}.

To address spatial sparsity, statistical methods such as Stratified Kriging (STK)~\cite{STK} and the Generalized Heterogeneity Model (GHM)~\cite{GHM} utilise spatial stratified heterogeneity (SSH) to divide complex regions into internally homogeneous sub-regions (strata). SSH describes how stratifying factors influence variable distribution, creating high intra-stratum homogeneity and inter-stratum heterogeneity. Ordinary Kriging is then applied within each stratum, relaxing the global Gaussian assumption.
However, these methods depend heavily on strata while overlooking their relationships, making interpolation dependent on individual strata and constrained by manually defined parameters. The coarse-grained strata framework also limits fine-grained kriging. Recently, inductive DL-based kriging has emerged as a more suitable solution for spatially sparse settings, as it enables inference at unseen locations without retraining~\cite{KITS,IGNNK}, unlike transductive methods~\cite{KCN}. However, both inductive and transductive frameworks typically rely on pairwise location correlations, lacking explicit mechanisms to encode region-level structures. This limits their ability to capture broader spatial semantics, such as regional coherence, and hinders accurate inference in sparsely observed areas.
%
%
%
A typical inductive kriging approach is KITS~\cite{KITS}, which randomly incorporates a large number of interpolated pseudo nodes into the training process as shown in Figure~\ref{fig:intro_fig} (c) may propagate interpolation errors and overlook SSH due to random generation, while also incurring additional computational cost. Besides, most DL methods~\cite{STK,KITS} fail to handle incomplete features across locations. A na\"ive solution is to process features separately (See SSIN~\cite{SSIN} in Figure~\ref{fig:intro_fig}(b). However, this method further reduces the set of usable locations in training, resulting with information loss and degraded performance in sparse datasets~\cite{IGNNK,SSIN,STGNP}.

In response to the limitation of current methods, we present an \underline{Anchor}-based Incremental and Stratified \underline{G}raph Learning Framework for Inductive Spatio-Temporal \underline{K}riging (AnchorGK). 

To mitigate the problem caused by sparsity and incomplete features, we propose an incremental Stratified Spatial Correlation Component (SSCC). 
Unlike decremental methods that only employ locations with the target feature for kriging, we take an \emph{incremental learning strategy} by utilizing the information from highly correlated features across locations, as illustrated in Figure~\ref{fig:intro_fig}~(d). Specifically, SSCC first identifies anchor locations, i.e., locations with the most available features, and establishes the stratum of each anchor location inspired by spatial autocorrelation theory and SSH~\cite{SpaAuto, SSH}. Each stratum is the largest polygon formed by an anchor location and the location closely relevant to the anchor, indicating the area governed by those locations. 
Subsequently, SSCC partitions the strata into finer-grained grid cells, referred to as sub-regions, to enable fine-grained estimation. SSCC employs a spatial correlation estimator to capture the correlations between known locations and unobserved sub-regions.
By employing both local and global kriging based on the estimated correlations, SSCC enables an initial interpolation of all features at strata. 



Building upon the anchor location-based spatial stratification framework, 
we introduce a dual-view Graph Learning Layer (GLL) comprising two key components -- (1) Cross-Feature Estimator (CFE) and (2) Cross-Strata Estimator (CSE). Specifically, the Cross-Feature Estimator employs a Graph Convolutional Network to obtain a localized feature representation in each strata. Following this, we employ a Kalman Filter Estimator to facilitate cross-feature fusion of all features. 
Since the Cross-Feature Estimator only captures specific local neighbors of the target unknown locations, we further design a Cross-Strata Estimator (CSE) to combine the information from all strata. Since each stratum independently represents a feature space specific to the corresponding anchor location, analogous to an ``expert'' in kriging, we design the Cross-Strata Estimator as a Mixture of Experts (MOE) to adaptively integrate representations obtained from strata. In this way, the dual-view graph learning layer integrates the diverse patterns across both features and locations.


Our contributions can be summarized as follows: (1) We propose a sub-region spatial correlation component to enable the modeling of correlations between known locations and unknown sub-regions in each stratum, mitigating the problem of sparse spatial distribution and missing features; 
(2) We propose a dual-view graph learning layer to obtain the representation of the target feature based on cross-feature and cross-strata views, capturing diverse patterns among features and locations; 
(3) We propose AnchorGK and conduct extensive experiments on three real-world datasets to show that AnchorGK outperforms the state-of-the-art (SOTA) models. 


\vspace{-1.5mm}  
\section{Related Work}
 

Kriging refers to a geostatistical technique for estimating the value of a variable at an unknown location based on observed values at known locations. It has been widely applied in various domains, leveraging spatial relationships for accurate predictions. 
Traditional statistical kriging methods, such as basic kriging and IDW~\cite{kriging1990,adaptiveIDW}, are often constrained by Gaussian assumptions and strict distance-based hypotheses, limiting their applicability in complex settings. To address these limitations, deep learning-based kriging has emerged, typically categorised into two main types: \textbf{Inductive vs. Transductive} models and \textbf{Multivariate vs. Univariate} models~\cite{IGNNK,KCN,SSIN,INCREASE,STGNP}.

Transductive kriging methods, such as Matrix Factorization and KCN, operate under the assumption that all nodes are observed during the training phase~\cite{GLTL,KCN}. These methods do not learn representations for unseen locations, making them unsuitable for dynamic or evolving datasets~\cite{KITS}. This limitation underscores the necessity of inductive kriging, which can adapt to new locations without retraining~\cite{IGNNK}. Prominent inductive kriging methods include IGNNK~\cite{IGNNK}, INCREASE~\cite{INCREASE}, and SSIN~\cite{SSIN}, which address unseen locations by either simulating their presence during training or dynamically generating their representations. Inductive kriging methods can be further divided into \textbf{decremental} and \textbf{incremental} approaches.
Decremental methods simulate unobserved locations by masking observed ones but struggle when node sparsity is high Incremental methods, like KITS~\cite{KITS}, enhance representation capabilities by generating additional pseudo-locations~\cite{KITS}. 
However, KITS overlooks spatial heterogeneity across regions, an assumption often invalid for large-scale, complex surfaces~\cite{GHM,STK}. Additionally, the introduction of pseudo-nodes increases computational cost and may propagate errors during the training.
Recent statistical methods, such as STK and GHM~\cite{STK,GHM}, divide regions into multiple strata to estimate target values within each stratum~\cite{GHM}. Nevertheless, these methods require manually defined parameters, making it computationally expensive for large-scale and diverse applications.

Feature-wise, most existing kriging methods, including IGNNK\\~\cite{IGNNK} and SSIN~\cite{SSIN},  focus on univariate spatial interpolation. While effective, these methods fail to fully capture cross-feature correlations~\cite{STGNP}. Currently, multivariate kriging has been explored by STGNP~\cite{STGNP} and STFNN~\cite{STFNN}. However, these models assume that the features are fully observed at all locations, restricting their ability to utilize the incomplete features across locations. Table~\ref{tab:related_work_transposed} summarises the differences between existing methods. 

\vspace{-1.5mm}  

\begin{table}[h!]
\scriptsize
\centering
\caption{Comparison of Models on the Support for Strata, Inductive Learning (Inductive), Incremental Training (Incremental), Spatially Sparsity Observations (Sparsity), Incomplete Observations (Incomplete), and Multivariate (Multivariate) Settings.}
\label{tab:related_work_transposed}
\renewcommand{\arraystretch}{0.9} 
\setlength{\tabcolsep}{1.5pt}  

\begin{tabular}{lcccccccccc}
\toprule
\textbf{Setting} & \textbf{OK} & \textbf{IDW} & \textbf{GHM} & \textbf{KCN} & \textbf{IGNNK} & \textbf{INCREASE} & \textbf{STGNP} & \textbf{KITS} & \textbf{SSIN} & \textbf{AnchorGK} \\
\midrule
Strata     & \xmark & \xmark & \cmark & \xmark & \xmark & \xmark & \xmark & \xmark & \xmark & \cmark \\
Inductive      & \xmark & \xmark & \xmark & \xmark & \cmark & \cmark & \cmark & \cmark & \cmark & \cmark \\
Incremental      & \xmark & \xmark & \xmark & \xmark & \xmark & \xmark & \xmark & \cmark & \xmark & \cmark \\
Sparsity       & \xmark & \xmark & \cmark & \xmark & \xmark & \xmark & \xmark & \cmark & \xmark & \cmark \\
Incomplete    & \xmark & \xmark & \xmark & \xmark & \xmark & \xmark & \xmark & \xmark & \xmark & \cmark \\
Multivariate    & \xmark & \xmark & \xmark & \xmark & \xmark & \xmark & \cmark & \xmark & \xmark & \cmark \\
\bottomrule
\end{tabular}
\end{table}
\vspace{-1.5mm}

\section{Methodology}

\subsection{Problem Definition}
Let the number of observed locations and unobserved locations be $N$ and $M$, respectively. The data with all $N$ observed locations is denoted by $\mathcal{X}=(X_{1},X_{2},...,X_{N})\in \mathbb{R}^{N\times T\times F}$, where $T$ is the length of time window and $F$ refers to the number of features. For the $i$-th location, $X_{i}=(x_{i,1},x_{i,2},...,x_{i,F}) \in \mathbb{R}^{T\times F}$ denotes its time series. Note that $x_{i,f}\in \mathbb{R}^{T}$ represents the time series for the $f$-th feature at the $i$-th location; if the $f$-th feature is unavailable, the corresponding series is set to zero, resulting in incomplete observed features. The value of $x_{i,f}$ at time $t$ is denoted by $x_{i,f}^{t}$. The goal of inductive spatio-temporal kriging is to estimate the unobserved time series at $M$ locations as $\mathcal{Y}=(\hat{Y}_1,\hat{Y}_{2},...,\hat{Y}_{M})\in \mathbb{R}^{M\times T\times F}$, where $\hat{Y}_{m} \in \mathbb{R}^{T\times F}$ is the estimated time series for the $m$-th unobserved location. 



\vspace{-1.5mm}  
\begin{figure*}[h]
\centering
\includegraphics[width=0.8\textwidth]{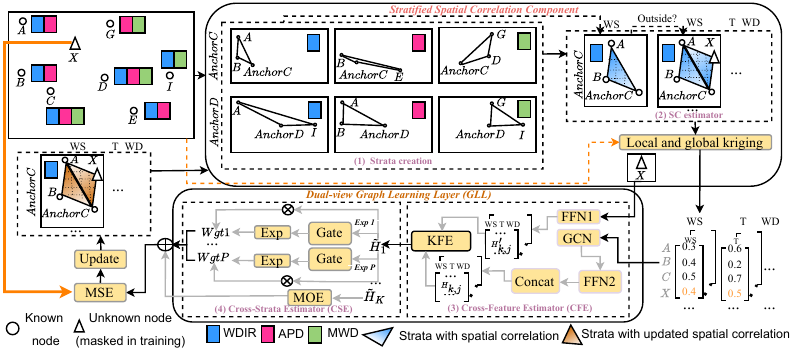}
\caption{The AnchorGK framework.}
\label{fig:Arch_MOESTUKF}
\end{figure*}
\vspace{-1mm}  

\subsection{Overall framework}

The architecture of AnchorGK is shown in Figure~\ref{fig:Arch_MOESTUKF}. 
To estimate $\mathcal{Y}$, AnchorGK employs two novel components: (1) Stratified Spatial Correlation Component (SSCC), which establishes the fine-grained estimation of spatial correlations between known locations and unknown locations; and (2) Dual-view Graph Learning Layer (GLL), which integrates the information from relevant features and locations for spatio-temporal kriging.


\subsection{Stratified Spatial Correlation Component}
\label{sec:SSCC}

Identifying the observed locations most related to the unobserved locations is the main challenge in spatio-temporal kriging~\cite{INCREASE}. While existing work has utilized auxiliary features to enhance performance~\cite{DGNN}, we aim to utilize all features available in the data. Specifically, we refer to locations that possess the majority of features as anchor locations and use them as references for estimating the features at any location in the surrounding areas. Consequently, we design a \textit{Stratified Spatial Correlation Component (SSCC)} to generate coarse-grained strata, fine-grained grid cells, and spatial correlation for fine-grained grid cells. 
The stratum is the maximum polygon defined by the anchor and its correlated neighbor locations. To obtain a fine-grained region representation, we further divide the stratum regions into smaller, evenly distributed grid cells \textit{Strata creation}. 
Subsequently, the \textit{Spatial Correlation Estimator}  constructs adjacency matrices to represent the spatial correlations between fine-grained grid cell and known locations. 
Considering unknown locations may not appear in a given strata, the SC estimator further updates the spatial correlations 
for the unknown locations outside the strata. The \textit{Local and Global Kriging} component provides an  initial feature estimation at all unknown locations as input to the Dual-view Graph Learning Layer.


\subsubsection{Strata creation}\label{sec:Strata_creation} 

Strata are spatially distinct regions which are often used to enhance the precision of spatial analysis~\cite{KED,GHM,PMSN}. Traditional methods for strata construction include Voronoi diagrams, geographically optimal zones-based heterogeneity (GOZH), and grid-based approaches~\cite{GHM,gridkriging}.
Voronoi diagrams partition space into regions where each point is closest to a designated generator, relying solely on proximity~\cite{voronoi}. In contrast, GOZH optimizes zones by considering intra-strata characteristics~\cite{GHM}.




In this work, we propose a new stratification method based on anchor locations that include most observed features. This is motivated by point pattern analysis, where anchor locations serve as a reliable reference point to ensure accurate detection of clustering, dispersion and randomness~\cite{pointtheory}. Based on each anchor location, our method establishes the strata by maximizing the intra-strata homogeneity. Unlike Voronoi and GOZH, which requires pair-wise computation, our approach can achieve $O(QN)$ runtime complexity, where $Q$ is the number of anchor locations. In practice, $Q$ is selected to be significantly smaller than $N$. Under this condition, the computational cost scales almost linearly with $N$.
To generate strata, we first collect a set of relevant locations for each anchor location by Pearson correlation considering the intra-strata homogeneity~\cite{SSH}.
%
%
Formally, 
given the $k$-th anchor location and the $f$-th feature, we obtain the set of relevant location IDs $S_{k,f}$ by:

\begin{equation}
S_{k,f} = \left\{ i \,\middle|\, x_{i,f} \in S_f \text{ and } i \in \text{Top}K\left(\text{Pearson}(x_{k,f}, x_{i,f})\right) \right\}
\label{eq:location_list}
\end{equation}                                  
where $x_{i,f}$ represents values of the $f$-th feature at the $i$-th location and $S_{f}$ represents the set of locations with the $f$-th feature available. We select the Top-$K$ node sets, where each corresponds to a node ID, such that the $K$ sets with $U$ nodes within each stratum exhibit the highest correlations.

\begin{figure}[htp] 
\centering
\begin{subfigure}{0.23\textwidth}
    \centering
    \includegraphics[width=\textwidth]{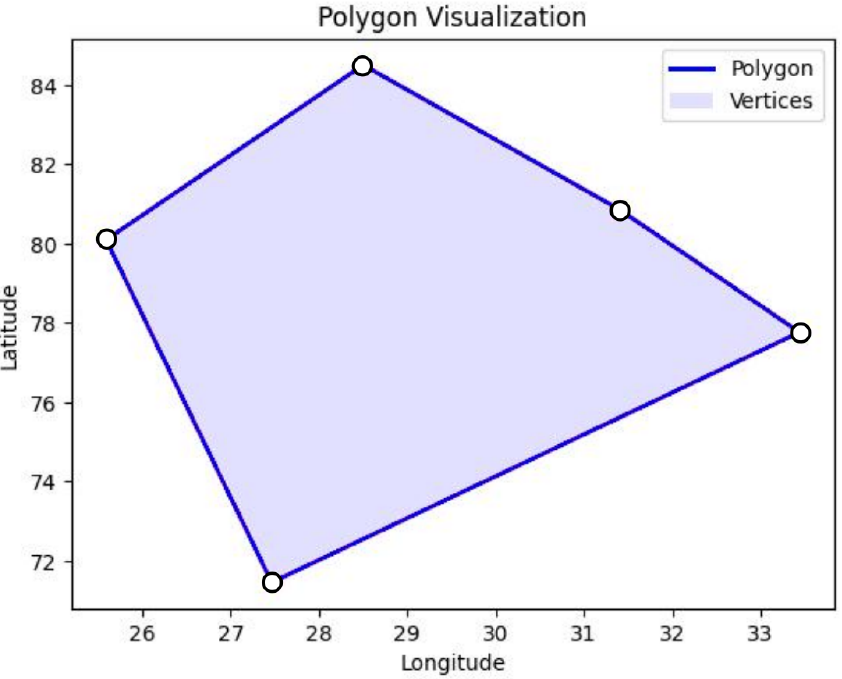}
    \subcaption*{(a) Strata}
\label{fig:coarse_poly}
\end{subfigure}
\hfill
\begin{subfigure}{0.23\textwidth}
    \centering
    \includegraphics[width=\textwidth]{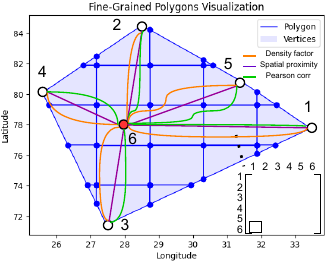}
    \subcaption*{(b) Grid cells}
    \label{fig:fine_poly}
\end{subfigure}
\vspace{-1em}
\caption{Coarse-grained strata and fine-grained grid cells.}
\label{fig:poly_vis}
\vspace{-1em}
\end{figure}

Based on $S_{k,f}$, we obtain the strata $P_{k,f}$ by connecting each location and obtaining the largest polygon as Figure~\ref{fig:poly_vis} (a) shows. Assuming a location can be uniquely described by geographic coordinates (i.e. latitude and longitude), let $L_{k,f}$ denotes the set of these coordinates for all locations in $S_{k,f}$. Following the Graham's scan~\cite{largeconvex}, we use the convex function to construct the convex strata:
$P_{k,f}=Convex(L_{k,f}),$
where $Convex$ refers to Graham's scan of finding the convex hull of a finite set of points in the plane with time complexity $O(SlogS)$. 

As the coarse-grained strata are imprecise of spatial correlations for unknown locations, we provide a fine-grained grid-based representation of grid cells following SpaBERT~\cite{spabert}, as illustrated in Figure~\ref{fig:poly_vis} (b). We define the fine-grained grid cells as: 
\begin{equation}
 P_{k,f}^{fine}=({P_{k,f}^{fine,1},P_{k,f}^{fine,2},...,P_{k,f}^{fine,v}}),
 \label{eq:fine_region}
\end{equation}
where $v$ represents the number of fine-grained grid cells in the strata. 

\subsubsection{Spatial Correlation Estimator}\label{sec:adj_rep}

After stratification, each grid cell represents an unobserved sub-region may contain unknown locations. To establish relationships between each grid cell
and the surrounding known locations, enabling unknown-to-known
estimations, we develop a spatial correlation estimator.
The estimated correlations are represented by an adjacency matrix, which is used with GNNs for subsequent estimations for unknown locations. Modeling spatial correlations is complex due to the intricate region-wise driven by spatial stratified heterogeneity~\cite{SSH}. Existing methods like kriging often fall short in addressing this complexity, especially when unknown regions dynamically interact with known locations~\cite{GHM,DAMR}.
To construct unified spatial correlation representations, we consider three key factors: Pearson correlation~\cite{INCREASE}, distance~\cite{DAMR}, and density~\cite{adaptiveIDW}.

\textbf{Density factor}:
We absorb the density factor in the construction of the adjacency matrix according to the well-established concepts in
point pattern analysis~\cite{pointtheory}.
According to the theory, absorbing neighbors' observations should be determined by
comparing the observed average nearest neighbor distance with the expected nearest neighbor distance. If the observed average nearest neighbor distance is greater
than the expected one, then the observed
point pattern is more dispersed than a random pattern. Motivated by previous work~\cite{adaptiveIDW}, we determine the density factor by the following equation:
\begin{equation}
\alpha= \frac{r_{obs}}{r_{exp}}, \quad r_{obs}=\sum{d_{m,n}}, \quad r_{exp}=\frac{1}{2(\frac{n}{\omega})^{\frac{1}{2}}},
\label{eq:den_factor}
\end{equation}
where $r_{obs}$ represents the observed average nearest neighbor
distance from target location to other nodes, $r_{exp}$ represents the expected nearest neighbor distance. $d_{m,n}$ stands for the distance between unobserved location $m$ and observed location $n$, $m$ and $n$ come from the unknown location $M$ and known location in $N$ respectively. $\omega$ stands for the area in the strata. The summary of variables can be found in Appendix~\ref{sec:Tab_of_variables}.

\textbf{Spatial proximity and Pearson correlations}:
The spatial proximity~\cite{grin} can be modeled by Eq.~\ref{eq:dist}:

\begin{equation}
\label{eq:dist}
{a}_{i,j}^{Dist}=\left\{\begin{array}{l}
\exp \left(-\frac{D_{i j}^{2}}{\sigma^{2}}\right), i \neq o \text { and } \exp \left(-\frac{D_{i j}^{2}}{\sigma^{2}}\right) \geq \epsilon \\
0, \text { otherwise. }
\end{array}\right.
\end{equation}

In Eq.~\ref{eq:dist}, ${D_{i j}}$
denotes the geographic distance between the $i$-th and $j$-th nodes,
$\sigma$ and $\epsilon$ are thresholds to control the distribution and sparsity of the adjacency matrix. As the distance-based adjacency matrix fail to support high spatial correlations between two remote locations, INCREASE\cite{INCREASE} measures the spatial correlation between $i$-th and $j$-th node on a $f$-th feature by Pearson correlation coefficient $\rho$.

\begin{equation}
\rho_{i,j}^{f}=max(0, Pearson(x_{i,f},x_{j,f})),
\end{equation}
where $Pearson$ represents the Pearson correlation coefficient function.

\textbf{Unified representation}:
We provide a unified adjacency matrix representation for the fine-grained known locations and fine-grained grid cells in each stratum. We represent each fine-grained grid cell by its center point. Specifically, the adjacency matrix of spatial correlations in a stratum is defined as follows:

\begin{align}
a_{i,j}^{f} &= \left(Pearson(x_{i,f},x_{j,f}) / e^{\frac{\lambda}{dist(i,j)}}\right)^{\rho_{i,j}^{f}}, \notag \\
a_{i,v}^{f} &= \sum_{j}^{U+1}\left(Pearson(x_{i,f},x_{j,f}) \times e^{\frac{\lambda}{dist(j,v)}}/e^{\frac{\lambda}{dist(i,j)}}\right)^{\rho_{i,j}^{f}}, \label{eq:uni_manner}
\end{align}
where $a_{i,j}^{f}$ represents the correlations between known locations, $a_{i,v}^{f}$ represents the correlations between known location and unknown location. $i,j \in N$ represents the $i,j$-th observed location, $v \in V$ represents the center point in the strata.
$\lambda$ represents the coefficient of distance-based weighting mechanism. $\alpha$ stands for the density factor in Eq.~\ref{eq:den_factor}. Therefore, we generate a unique adjacency matrix for each fine-grained grid cells. Following this way on all elements in the overall adjacency matrix, we obtain the $A_{k,v}^{w} \in \mathbb{R}^{(U+2)\times(U+2)}$.

\textbf{Target location outside the sub-region}:
Given an unknown location, we want to speculate the most possible sub-region it belongs to choose the most relevant SC for graph message passing. 
If the unknown location is not located in any coarse-grained strata, we will then expand the largest convex polygon obtained in  Section~\ref{sec:Strata_creation} to include the unknown location. Subsequently, we follow Eq.~\ref{eq:fine_region} to divide it into fine-grained grid cells and establish the spatial correlations among all obtained sub-regions following Eq.~\ref{eq:uni_manner}. To this end, even if the unknown location is outside the strata of an anchor location, we can determine its correlation with the known locations.

We update the parameters in SSCC following Markov chain Monte Carlo (MCMC)~\cite{MCMC} to capture the dynamics of spatial correlations by exploring the high-dimensional posterior space of parameters in SSCC.

\subsection{Local and Global Kriging}

The global kriging utilizes all existing features among all locations, while local kriging only utilises existing features in each stratum. 

\textbf{Local Kriging}:
On the $k$-th anchor location and the $f$-th feature, the interpolated value on the $m$-th unknown location following Kriging method will be:

\begin{equation}
\overline{x}_{m,f}=Kriging(x_{1,f},x_{2,f},...,x_{u,f},x_{k,f}), 
\end{equation}
where $\overline{x}_{m,f}$ stands for interpolated value on the $m$-th unknown location. For convenience, we combine interpolated value, anchor location value and neighbors in anchor locations together as:

\begin{equation}
x_{k,f}^{aug}=[x_{1,f},x_{2,f},...,x_{U,f},x_{k,f},\overline{x}_{m,f}],    
\end{equation}
where $x_{k,f}^{aug}\in \mathbb{R}^{T\times(U+2)}$.

\textbf{Global Kriging}:
The global kriging utilizes all existing features across all locations for $m$-th unknown location. 

\begin{equation}
\overline{x}_{f}^{global}=Kriging(x_{1,f},x_{2,f},...,x_{V,f}), \quad \overline{x}_{f}^{global}\in \mathbb{R}^{T\times f},
\end{equation}
where $V$ represents the number of known features on the $f$-th feature.

\subsection{Dual-view Graph Learning layer}\label{sec:msg_passing}

Building on the implementation of local and global kriging, we obtained an initial estimation for all features, inspired by DAMR~\cite{DAMR}. However, this estimation lacks precision as it overlooks the correlations between cross-feature and cross-location relationships. To address this limitation and uncover diverse patterns among features and locations, we propose a dual-view Graph Learning Layer (GLL), comprising two key components: the Cross-Feature Estimator (CFE) and the Cross-Anchor Estimator (CAE).


\subsubsection{Cross-Feature Estimator (CFE)}
Given $x_{k,f}^{aug}$, $\overline{x}_{f}^{global}$, and $A_{k,v}^{w}$ for each stratum, we apply GCN and KFE for message passing.
We utilize GCNs to learn the representation on $U+2$ locations: anchor locations, neighboring locations and unknown locations:

\begin{equation}
H^{\prime}_{k,f}=\hat{D}_{k,f}^{- 1 / 2} \hat{A}_{k,f}^{w} \hat{D}_{k,f}^{- 1 / 2} X_{k,f}^{aug} W_{k,f} + b_{k,f},
\end{equation}
where $H^{\prime}_{k,f} \in \mathbb{R}^{(U+2)\times F^{\prime}}$ is the latent representation of GCN, $F^{\prime}$ is the number of output channels in GCN. $\hat{A}_{k,f}^{w}=A_{k,f}^{w}+I$ is the adjacency matrix with the self-loops and $I$ stands for the identity matrix, $\hat{D}$ is the degree matrix.

\textbf{Kalman Filter Estimator (KFE)}:
We employ a Kalman Filter Estimator (KFE) to boost the cross-feature fusion among all features.


Formally, we define the state vector as follows: \\
$H_{t}^{m}=(H_{t,1}^{m},H_{t,2}^{m},...,H_{t,F}^{m}),$
where $H_{t}^{m}\in \mathbb{R}^{F^{\prime}\times T \times j}$ represent the overall hidden state on all features. The measurement for all features are constructed by $\overline{x}^{global} \in \mathbb{R}^{T\times j}$. We choose the Unscented Kalman Filter as the basic filter in KFE due to its high precision among Kalman Filters and the low reliance on the linear state transition and measurement functions compared to traditional KF~\cite{UKF}:
\begin{equation}
\tilde{H}_{k}^{m}=KFE(FFN1(H_{k}^{m}),FFN2(\overline{X}^{global})).
\label{eq:cross_feature}
\end{equation}

Specifically, FFN1 and FFN2 serve as state transition functions and measurement functions. We apply FFN1 on each timestep and map from the $H^{m}_{k} \in \mathbb{R}^{F^{\prime}\times T\times j}$ to $H^{m,output}_{k} \in \mathbb{R}^{ T\times j}$, we apply FFN2 on unknown location after the global kriging interpolation. The state vector is constructed based on the number of features $F$. The CFE algorithm is detailed in Appendix~\ref{alg:CFE}.

\subsubsection{Cross-Strata Estimator (CSE)}

The cross-anchor message passing utilises a Mixture-of-Experts (MOE) framework to fuse the outputs of anchor-based strata. To achieve this, we propose a Cross-Strata Estimator (CSE), where each anchor location independently represents a feature space unique to its strata, akin to an "expert" in kriging. This structure motivates the adoption of MOE for adaptively integrating the diverse representations from anchor locations, facilitating the optimal fusion of spatial information.

The fusion process is defined as:

\begin{equation}
   \tilde{H}_{k}=MOE( [\tilde{H}_{1}^{m},\tilde{H}_{2}^{m},...,\tilde{H}_{K}^{m}]).
\end{equation}

The MOE process can be formulated as:
\begin{equation}
\mathcal{Y}=
\sum_{k=1}^{K}\sum_{p=1}^{P}G_{p}(\widetilde{H}_{k})E_{p}(\widetilde{H}_{k}),    
\label{eq:MOE_detail}
\end{equation}
where the $p$-th gate component $G_{p}$ decides which experts (E) to send a part of input, the gate component is based on the softmax function~\cite{MOE}. The MOE adaptively integrates spatial information by dynamically assigning input features to the most suitable anchor representations via a gating mechanism. This enhances flexibility, reduces redundancy, and promotes specialization, enabling efficient fusion of diverse spatial features and improved performance.

\subsection{Training and Testing}

In the training stage, 
the $m$-th unobserved location is obtained from set of $M$ unobserved locations, the latter  obtained by masking entire locations.
We utilize all of known locations $N$ to infer the values of $m$-th location and repeat this for all $M$ locations~\cite{IGNNK}. We utilize the RMSE between the estimated value $\hat{Y}_{m}$ and true value $Y_{m}$. Once trained, the trained AnchorGK can be used to test for any locations without observations. We provide a detailed training and inference figure to illustrate the inductive kriging fashion in Figure~\ref{fig:train_strategy}.

\subsection{Complexity analysis}

The Overall Complexity of AnchorGK is written in Eq.~\ref{eq:timecomp}:

\begin{equation}
O\big(Q \cdot U \log U + Q \cdot U^2 +Q \cdot F^2 \cdot T + Q\cdot U \cdot G ).
\label{eq:timecomp}
\end{equation}

The overall complexity of AnchorGK combines contributions from its key components. The complexity expression $Q \cdot (U \log U)$ reflects the computational cost of performing a sorting operation (e.g., Top-$K$) over the neighbour set $U$ for all $Q$ anchor nodes. 
The remaining terms capture the training complexities of the GLL and MoE modules. 
CFE exhibits a quadratic complexity with respect to the neiborhood $U$, the KFE module demonstrates quadratic complexity with respect to the feature dimension $F$, while the gating operation in the CSE module scales linearly with the number of selected gates $G$. Taken together, the use of anchor station processing significantly reduces the overall training time, especially when the number of selected anchor stations $Q$ remains relatively small.

\section{Experiments}
\label{sec:experiment}
In this section, we aim to answer the following questions:
\begin{itemize}
    \item \textbf{RQ1}: How does AnchorGK perform against SOTA imputation methods on real-world data sets?--Overall performance
    \item \textbf{RQ2}: What is the impact of each component of AnchorGK on its spatial interpolation performance?--Ablation study
    \item \textbf{RQ3}: Can the Stratified Spatial Correlation Component learn the correlated spatial correlations?--Case study
    \item \textbf{RQ4}: How sensitive is the model to the change of different parameters?--Parameter sensitivity analysis
    \item \textbf{RQ5}:  How does the performance of AnchorGK scale w.r.t. the number of
features and nodes in the network?--Spatial sparsity and feature availability analysis
\end{itemize}

\subsection{Experiment Setup}

\subsubsection{Data sets}
We conduct the experiment on three real-world datasets: 
Unite States Coastline (UScoast)~\cite{NDBC,NDBC2}\footnote{https://www.ndbc.noaa.gov/obs.shtml}, Shenzhen~\cite{SZ_source,DAMR} and Te Hiku dataset~\cite{INCREASE,soil_moisture} . We select these datasets considering their sparsity and varying available features. Table~\ref{tab:Summary_stat} shows the statistics of these datasets.

\begin{table}[!ht]
\centering
\caption{Summary of three data sets }
\vspace{-1em}
\label{tab:Summary_stat}
\begin{tabular}{ccccc}
\hline
& UScoast        & Shenzhen          & Te Hiku             \\ \hline
Feature     & 9         & 6           & 4                  \\
Node       & 103        & 10           & 24              \\
Timesteps  & 8784    & 7296      & 76400             \\
Granularity & 1 hour  & 1 hour   & 15 minutes         
 \\ \hline
\end{tabular}
\vspace{-1em}
\end{table}


\textbf{UScoast}
The UScoast dataset is provided by the National Data Buoy Center (NDBC), consisting of hourly climate observations from approximately 90 buoys and 60 Coastal Marine Automated Network (C-MAN) stations~\cite{NDBC,NDBC2}. It spans a longitude range of \(46.327^\circ\text{W}\) to \(177.703^\circ\text{W}\) and a latitude range of \(14.453^\circ\text{N}\) to \(57.061^\circ\text{N}\). The dataset is highly sparse and exhibits significant feature imbalance: only 35 locations have complete features; the rest show varying levels of missingness.

\textbf{Te Hiku}
Similar to prior hydrological studies~\cite{INCREASE,soil_moisture}, we use soil moisture data from the Te Hiku forest region. It covers 24 locations within longitude $173.104^\circ\mathrm{W}$ to $173.149^\circ\mathrm{W}$ and latitude  $34.905^\circ\mathrm{S}$ to $34.933^\circ\mathrm{S}$. Unlike UScoast, this dataset is dense and balanced, with all locations having complete feature sets.

\textbf{Shenzhen}
The Shenzhen dataset\footnote{https://www.microsoft.com/en-us/research/project/City\_computing/} comes from prior work~\cite{SZ_source,DAMR} and comprises 10 monitoring stations with 6 air-quality features. It spans from 2014 to 2015 in Shenzhen, China, covering a latitude range of \(22.491^\circ\text{N}\) to \(22.72^\circ\text{N}\) and a longitude range of \(113.946^\circ\text{E}\) to \(114.497^\circ\text{E}\). Similar to other air quality datasets~\cite{STGNP}, the sensors are primarily located in urban or suburban areas. Like Te Hiku, this dataset is dense and exhibits no feature imbalance.

\subsubsection{Baselines}




We select baselines from both statistical and deep learning (DL)-based approaches. Specifically, the statistical methods are categorized into \textbf{strata-based} and \textbf{non-strata-based} statistical approaches, while the DL methods are classified into \textbf{incremental} and \textbf{decremental} approaches. It is important to note that DL methods do not include strata-based approaches, and statistical methods do not include incremental approaches.

\textbf{Non-strata-based statistical approaches}:
\begin{itemize}
    \item OK: Ordinary Kriging~\cite{ok}, which models spatial correlation as an exponential decay with squared distance.
    \item IDW: Inverse distance weighting~\cite{KrigingReview}, a deterministic interpolation method based on distance-weighted averaging.
\end{itemize}

\textbf{Strata-based statistical approaches}:

\begin{itemize}
    \item \textbf{GHM}: Generalized Heterogeneity Model~\cite{GHM} that partitions the area into strata based on sample values and locations.
\end{itemize}

\textbf{Decremental DL methods}:
\begin{itemize}
    \item KCN: Kriging Convolutional Network~\cite{KCN}, aggregating information from K-nearest neighbors for kriging.
    \item IGNNK: Inductive Graph Neural Network for spatio-temporal kriging~\cite{IGNNK}, leveraging spatio-temporal information from all nodes via distance-based adjacency matrices.
    \item SSIN: Self-Supervised Learning for Rainfall Spatial Interpolation~\cite{SSIN}, using SpaFormer to model spatial correlations.
    \item INCREASE: Inductive Graph Representation Learning for Spatio-Temporal Kriging~\cite{INCREASE}, encoding heterogeneous spatial relations to capture dependencies across locations.
    \item STGNP: Spatio-Temporal Graph Neural Processes~\cite{STGNP}, jointly modeling uncertainty and spatio-temporal correlations by causal convolutions and cross-set graph networks.
\end{itemize}

\textbf{Incremental DL methods}:
\begin{itemize}
    \item KITS: Kriging with Incremental Training Strategy~\cite{KITS}, introducing virtual nodes during training to mitigate sparsity.
\end{itemize}

\subsubsection{Hyper-parameter Settings}\label{sec:hepyer_settings}

For AnchorGK, the number of neighbors in each subgraph is set to $5$; The decay coefficient is tuned, assuming values in the range (0,1.6,0.1); The hidden dimension in MOE model is set to 16; The $\alpha, \beta $ and $\kappa$ are set to 0.1, 2.0 and 0.1, respectively in KFE. The hidden state in GCN is set to $16$. The kernel function is set to linear. All experiments were run on Intel (R) Core (TM) i9-13900HX CPU, 32GB RAM, and NVIDIA RTX 4080 GPU. We tune the hyperparameters by grid search on both AnchorGK and other baselines~\cite{Gridsearch}. All baselines are evaluated using the same unobserved locations and observed locations. 
On geostatistical methods, we set the power of 2 for IDW following the same baseline setting in \cite{SSIN}; On SSIN, the number of attention heads $H$ and hidden states $d_{k}$ in attention scheme are set to $2$ and $16$. The dimension of feed-forward network $d_{f}$ is set to 256. In IGNNK and INCREASE, the number of unknown location percentage and mask node during training are both set to $20\%$. In IGNNK~\cite{IGNNK}, the hidden state of GNN is set to 25, the diffusion convolution step is set to $1$, the number of max training episode is set to 750. On KCN~\cite{KCN}, the hidden size, kernel length and dropout rate are set to $20, 0.25$ and $1.05$, respectively. In GHM~\cite{GHM}, the number of strata is set to 3 following the default setting. The number of optimal neighbors $N_{max}$ is set to 16. In INCREASE~\cite{INCREASE}, $P$ is also set to 24 following the default setting. 
 In KITS~\cite{KITS}, the node insetion percentage is set to 50\%, $\lambda$ is set to 1 which controls the importance of pesudo labels. The decay coefficient is tuned by the range (0,1.6,0.1) to be consistent with AnchorGK. In STGNP~\cite{STGNP}, the number of STGNP layers is set to 1 and channel numbers $[8u, 16u, 32u]$ are set to 1.

\subsection{RQ1: Overall performance}

To address \textbf{RQ1}, we present the MAE and RMSE in Table~\ref{tab:overall_perf}. Note that, unlike univariate-based DL methods (e.g., IGNNK and SSIN), which calculate the differences using de-normalized results, we compute the MAE and RMSE between the normalized predicted and true values for multivariate results to eliminate the influence caused by dominated variable follwoing KCN~\cite{KCN} and GHM~\cite{GHM}. We have the following observations: \textbf{Firstly}, for strata-based methods, the GHM method demonstrates greater robustness compared to IDW and OK. While OK performs better on relatively dense datasets such as Te Hiku and Shenzhen, GHM exhibits more stable results on the sparse dataset UScoast, highlighting the advantage of stratified approaches for handling sparse data. \textbf{Secondly}, among decremental DL methods, SSIN outperforms other baselines due to its SpaFormer module, highlighting the great success in learning spatial correlations~\cite{KCN,IGNNK,STGNP}. \textbf{Thirdly}, results indicate that incremental training approach KITS outperforms most baselines across the majority of datasets, particularly excelling in Te Hiku dataset. \textbf{Fourthly}, AnchorGK outperforms all baselines through its utilization of SSCC and GLL motivated by strata and spatial correlation estimation~\cite{GHM,SSIN}, excelling in capturing fine-grained spatial correlations as well as cross-anchor and cross-feature correlations. 
\textbf{Finally}, the improvement of AnchorGK is statistically significant based on post-hoc Nemenyi test~\cite{Herbold2020}. As shown in Table~\ref{tbl:stat_results}, AnchorGK achieves the highest mean rank (MR), the lowest median (MED) and the lowest absolute value of Akinshin’s gamma $\gamma^\prime$~\cite{Akinshin}. 

\begin{table}[h]
\centering
\caption{Comparison on three datasets over 20 runs. Improvements are statistically significant based on the post-hoc Nemenyi test at \( \alpha = 0.05 \). Imp denotes improvement over the best baseline.}


\label{tab:overall_perf}
\setlength{\tabcolsep}{1.2mm}{
\begin{tabular}{lcc|cc|cc}
\hline
Datasets    & \multicolumn{2}{c}{UScoast} & \multicolumn{2}{c}{Te Hiku} & \multicolumn{2}{c}{Shenzhen} \\ \hline
Metrics     & MAE          & RMSE         & MAE          & RMSE        & MAE           & RMSE   \\ \hdashline
OK         & -            & -            & 0.068        & 0.094       & \underline{0.062}         & \underline{0.091}  \\
IDW         & 0.181        & 0.226        & 0.263        & 0.376       & 0.121         & 0.152  \\
GHM         & 0.164
        & 0.207
        & 0.179         &  0.213     & 0.153        & 0.179  \\

KCN         & 0.757        & 0.962        & 0.227        & 0.277       & 0.358         & 0.535  \\ 
IGNNK       & 0.348        & 0.393        & 0.234        & 0.355       & 0.116         & 0.148  \\
INCREASE       & 0.174        & 0.218        & 0.136        & 0.184       & 0.129         & 0.177  \\
STGNP       & 0.158        & 0.184        & 0.117        & 0.149       & 0.106         & 0.159  \\
KITS       & 0.117        & 0.146       & \underline{0.063}        & \underline{0.092}      & 0.089         & 0.112  \\
SSIN        & \underline{0.102}        & \underline{0.127}        & 0.072        & 0.110       & 0.074         & 0.096  \\
AnchorGK      & \textbf{0.084} & \textbf{0.114} & \textbf{0.058} & \textbf{0.082} & \textbf{0.046} & \textbf{0.067} \\ \hdashline
Imp & 17.6\%       & 10.2\%       & 7.9\%        & 10.9\%       & 25.8\%         & 26.4\%  \\ \hline
\end{tabular}}
\end{table}

\begin{table}[t]
\caption{Significant test of the algorithms}
\centering
\footnotesize
\setlength{\tabcolsep}{3pt} 
\begin{tabular}{@{}lrrrrrrr@{}}
\toprule
Metric & AnchorGK & SSIN & KITS & STGNP & INCREASE & IGNNK & KCN \\
\midrule
MR$\uparrow$              & \textbf{6.600} & 5.850 & 5.450 & 3.850 & 3.250 & 2.000 & 1.000 \\
MED$\downarrow$           & \textbf{0.086} & 0.105 & 0.118 & 0.155 & 0.173 & 0.352 & 0.761 \\
$\gamma^{\prime}\uparrow$ & \textbf{0.000} & -0.896 & -2.037 & -4.042 & -5.450 & -14.690 & -40.398 \\
\bottomrule
\end{tabular}
\label{tbl:stat_results}
\end{table}

\subsection{RQ2: Ablation study}

We also conduct the ablation study across three datasets to show the importance of Stratified Spatial Correlation Component (SSCC) and Graph Learning Layer (GLL) in AnchorGK. 

\begin{figure}
 \centering \includegraphics[width=\linewidth]{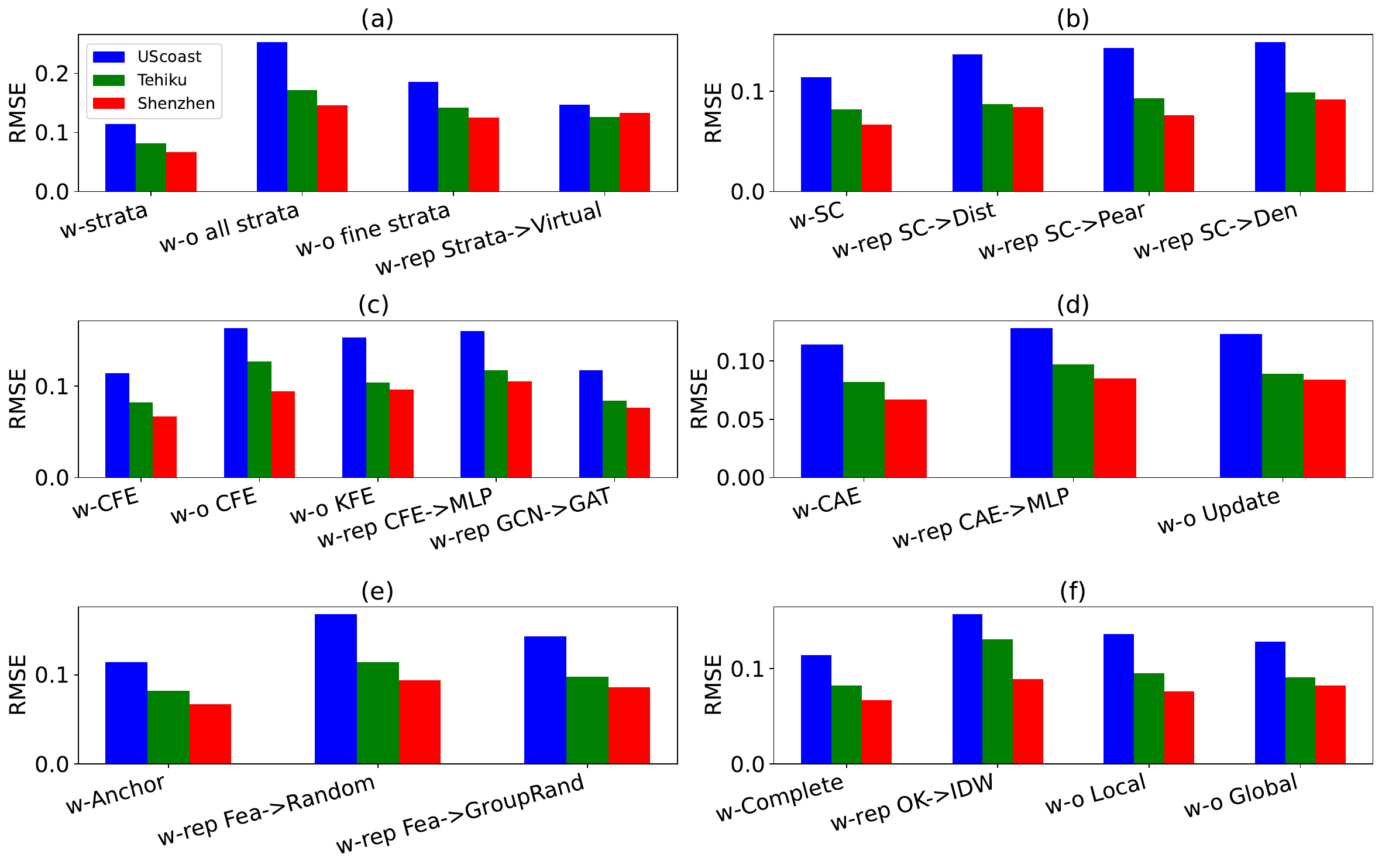}
    \caption{Ablation study on (a) Strata, (b) SC estimator, (c) CFE, (d) CAE, (e) Anchor selection strategy and (f) kriging strategy across three data sets.}
    \label{fig:ablation}
\end{figure}

\subsubsection{Impact of strata}

Figure~\ref{fig:ablation} (a) illustrates the necessity of utilizing strata. AnchorGK leverages multiple strata by dividing the graph based on different anchor locations. Compared to using strata (w-strata), treating all strata as a single graph (w-o strata) significantly deteriorates performance. Similarly, using only coarse-grained strata and neglecting the fine-grained strata (w-o fine strata) also negatively impacts the results. In addition, we replace the strata-based approach with the incremental training strategy of KITS~\cite{KITS} ($\text{w-rep Strata}\rightarrow\text{Virtual}$). Specifically, we adopt the Kriging Model and Node-aware Cycle Regulation (NCR) of KITS to insert pseudo nodes and their features. Since KITS does not create strata, we consider the whole graph as one large strata and pass it through the GLL layer.
The results demonstrate that utilizing fine-grained strata effectively enhances kriging performance and outperforms the pseudo node strategy.

\subsubsection{Impact of SC estimator}
Figure~\ref{fig:ablation} (b) highlights the importance of the SC estimator. We replaced the composite factors in Eq.~\ref{eq:uni_manner} with individual factors to evaluate their contributions: Spatial proximity($\text{w-rep SC}\rightarrow\text{Dist}$), Pearson correlation coefficient ($\text{w-rep SC}\rightarrow\text{Pear}$), and density-based factors ($\text{w-rep SC}\rightarrow\text{Den}$). The results show that distance is significant in Te Huku dataset.
Removing other factors while retaining only distance resulted in a 6.1\% $(0.082 \rightarrow 0.087)$ increase in RMSE. 
However, the Shenzhen dataset demonstrates that Pearson correlation coefficient is more critical than the other two factors. These findings underscore the Indispensability of the combined effects of all three factors in the SC estimator.


\subsubsection{Impact of CFE component}

We conduct the experiment by removing the Cross-Feature Estimator (w-o CFE) as well as the Kalman Filter Estimator (w-o KFE), replacing overall KFE by MLP (w-rep $\text{KFE} \rightarrow \text{MLP}$) and replacing GCN by GAT (w-rep GCN$ \rightarrow$ GAT).
Figure~\ref{fig:ablation} (c) illustrates that CFE is indispensable across three different datasets. Additionally, the choice of GCN and KFE is well-justified. Omitting the KFE component (w-o KFE) results in a performance degradation of 26.8\% ($0.082\rightarrow 0.104$) in the Te Hiku dataset. Similarly, replacing GCN with GAT (w-rep GCN$\rightarrow$ GAT) leads to a performance degradation in the Shenzhen dataset. These results underscore the importance of CFE component.




\subsubsection{Impact of CSE}
Figure~\ref{fig:ablation} shows that replacing CSE by MLP
($\text{w-rep CSE} \rightarrow \text{MLP}$)
will restrict the model to perform well in spatial interpolation. Similarly, updating the spatial correlations (w-o Update) in the SC estimator is essential. The use of MCMC has demonstrated its effectiveness by improving performance across all three datasets, confirming that MCMC is highly informative in updating the spatial correlations.

\subsubsection{Impact of anchor selection}\label{sec:anc_selection}
We compare our choice of anchor locations with two variants: (a) Random selection (w-rep Fea $\rightarrow$ Random) and (b) Uniform sampling by groups (w-rep Fea $\rightarrow$ Grouprand). The latter variant divides the locations into groups according to their available features and performs uniform sampling within each group. The results show that our choice of anchor locations outperforms other strategies.

\subsubsection{Impact of kriging strategy}\label{sec:kriging_strategy}
We also show that in the feature completeness method, using global and local kriging with ordinary kriging outperforms other approaches: with IDW (w-rep OK $\rightarrow$ IDW), without local kriging (w-o Local), and without global kriging (w-o Global).
 
\subsection{RQ3: Case study and visualization}

\subsubsection{Visualization of kriging results}

\begin{figure}
\centering   \includegraphics[width=0.8\linewidth]{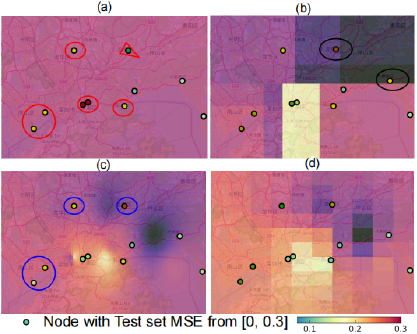}
\caption{Visualization on (a) Ordinary Kriging (b) Strata-based GHM model~\cite{GHM} (c) SSIN model~\cite{SSIN} (d) AnchorGK. }
\label{fig:two_visual}
\end{figure}

To better understand how AnchorGK behaves compared to other baseline models, we visualize the kriging results of the air quality index PM10 on Shenzhen dataset following the visualization process of SATCN~\cite{satcn}. Figure~\ref{fig:two_visual} shows that ordinary kriging captures global effects but omits local subtle changes, leading to poor performance in some regions (red circles), though beneficial in others (red triangles). GHM~\cite{GHM} partitions data into coarse-grained strata, ignoring cross-strata correlations (black circles).  SSIN~\cite{SSIN} balances global and local effects via attention mechanisms but struggles in sparse regions (blue circles). AnchorGK employs effective stratification with fine-grained sub-regions, thus enhancing the performance.

\subsubsection{Visualization of spatial correlation update}

\begin{figure}
    \centering
    \includegraphics[width=\linewidth]{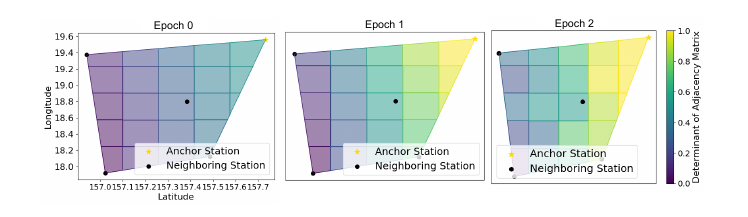}
    \caption{Correlation visualization on the first three epochs. }
    \label{fig:Visual_three}
\end{figure}

We conduct a case study on UScoast dataset to better illustrate the effectiveness of the training process. After the initialization of the fine-grained strata, each sub-region is represented by a unique adjacency matrix. To show the changes of adjacency matrices in different sub-regions, we calculate their determinant values~\cite{ADJdeterminant}. Figure~\ref{fig:Visual_three} illustrates distinct determinant values by colours in the first 3 epochs. 
Through the training process, 
certain sub-regions (denoted by yellow coloured cells) exhibit significant changes in epoch 1 and epoch 2. 


\subsection{RQ4: Sensitivity study}
We conduct sensitivity studies on the UScoast dataset.

\subsubsection{Impact of neighbors in each strata}
\begin{figure}
    \centering
    \includegraphics[width=\linewidth]{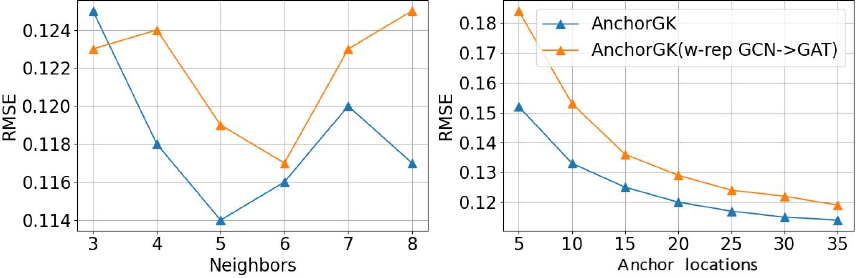}
    \caption{Sensitivity studies on neighbors and locations.}
    \label{fig:sensi}
    \vspace{-1em}
\end{figure}

To investigate how the number of neighbors $U$ influences the overall performance, we pick the number of neighbors in the range of $[3,8]$ and report the RMSE on the left side of Figure~\ref{fig:sensi}. We also test an alternative model by replacing GCN with GAT. The optimal $U$ of AnchorGK is 5, and the optimal $U$ of the alternative model with GAT is 6. We observed similar findings to those reported in INCREASE~\cite{INCREASE}, where an initial increase in the number of neighbors enhances results; however, further increases introduce noise, ultimately degrading performance.

\subsubsection{Impact of anchor locations in strata}

To evaluate the influence of the number of anchor locations on the performance,
 we test different numbers of anchor locations in $[5,10,15,20,25,30,35]$. As shown in Figure~\ref{fig:sensi}, introducing more anchor locations will improve the accuracy of the model.

\subsection{RQ5: Availability and sparsity experiment}

Table~\ref{tab:imbalancesparse} highlights the advantages of AnchorGK under varying spatial sparsity and feature availability. "Availability" refers to incomplete features, defined as the proportion of features randomly masked, while "Sparsity" denotes the proportion of sensors that are randomly masked~\cite{KITS}. The results show that OK performs the worst in both scenarios, reflecting its strong dependence on location density, consistent with the findings in~\cite{IGNNK,KITS}. While KITS~\cite{KITS} achieves comparable results with SSIN~\cite{SSIN}. Notably, AnchorGK excels in highly available (60\%) and sparse (40\%) scenarios, demonstrating its ability to establish strata via anchor locations and accurately estimate spatial correlations, even under extreme feature-masking conditions.


\begin{table}[h]
\centering
\caption{RMSE comparison across different missing data scenarios in Shenzhen data set. Results are averaged over 5 runs. Improvements (\%) are relative to the best baseline models.}
\label{tab:imbalancesparse}
\setlength{\tabcolsep}{0.8mm}{
\begin{tabular}{lc|ccc|ccccc}
\hline
\textbf{Metd} & \multicolumn{1}{c}{\textbf{Full}} & \multicolumn{3}{c}{\textbf{Availability}} & \multicolumn{3}{c}{\textbf{Sparsity}} \\ \cline{2-9}
 & \textbf{0\%} & \textbf{20\%} & \textbf{40\%} & \textbf{60\%} & \textbf{20\%} & \textbf{40\%} & \textbf{60\%} \\ \hline
OK & 0.094 & 0.127 & 0.148 & 0.155 & 0.185 & 0.228 & 0.359 \\
SSIN & 0.110 & 0.122 & 0.139 & \underline{0.146} & 0.153 & 0.166 & 0.198 \\
KITS & \underline{0.092} & \underline{0.118} & \underline{0.135} & 0.158 & \underline{0.141} & \underline{0.157} & \underline{0.174} \\
Ours & \textbf{0.082} & \textbf{0.098} & \textbf{0.114} & \textbf{0.128} & \textbf{0.125} & \textbf{0.132} & \textbf{0.147} \\ \hdashline
\textbf{Imp} & 10.9\% & 16.95\% & 15.56\% & 18.99\% & 11.35\% & 15.92\% & 15.52\% \\ \hline
\end{tabular}}
\vspace{-1em}
\end{table}

\subsection{RQ6: Running time comparison}
The runtime statistics and memory usage are reported in Table~\ref{tab:time_memory_compact}. While the pre-processing stage of our method incurs a slightly higher computational cost compared to existing approaches, the incorporation of the strata-based kriging framework significantly reduces training time across all three datasets. This observation aligns well with the theoretical time complexity analysis presented in Equation~\ref{eq:timecomp}, thereby validating the efficiency of our design.

\begin{table}[htbp]
\centering
\footnotesize
\caption{Pre-processing (Pre), training time (Train) in seconds, and estimated memory usage (Mem) in GB across datasets.}
\label{tab:time_memory_compact}
\setlength{\tabcolsep}{3.5pt}
\begin{tabular}{lccccccccc}
\toprule
\textbf{Model} & \multicolumn{3}{c}{\textbf{UScoast}} & \multicolumn{3}{c}{\textbf{Shenzhen}} & \multicolumn{3}{c}{\textbf{Te Hiku}} \\
\cmidrule(lr){2-4} \cmidrule(lr){5-7} \cmidrule(lr){8-10}
& Pre & Train & Mem & Pre & Train & Mem & Pre & Train & Mem \\
\midrule
STGNP     & 68.7  & 4357.6 & 10.3 & 10.6 & 896.4  & 0.57 & 48.8 & 2248.6 & 12.4 \\
KITS      & \textbf{35.4}  & 6879.3 & 14.7 & \textbf{7.3}  & 1127.8 & 0.86 & \textbf{27.6} & 4683.2 & 18.4 \\
SSIN      & 48.6  & 3749.5 & 5.2  & 8.5  & 684.5  & 0.43 & 33.1 & 1694.5 & 7.9 \\
AnchorGK  & 109.5 & \textbf{786.4} & \textbf{2.1} & 29.7 & \textbf{237.8} & \textbf{0.38} & 87.2 & \textbf{439.7} & \textbf{3.4} \\
\bottomrule
\end{tabular}
\vspace{-1em}
\end{table}

\section{Conclusion}
This paper addresses the challenges of spatio-temporal kriging posed by sparse location distribution and missing features at different locations. We introduce AnchorGK, a graph learning framework that leverages anchor locations to
create fine-grained grid cells within their strata and a dual-view graph learning layer that employs cross-feature and cross-anchor message-passing mechanisms
for spatio-temporal kriging. Extensive experiments on three real-world datasets show that AnchorGK outperforms the SOTA methods and reduces MAE up to 12.27\%. We will extend AnchorGK to support other tasks in the future work.

\begin{acks}
This research was supported by Shenzhen Science and Technology Program No. SYSPG20241211173609009. Additionally, the first author received support from the China Scholarship Council (CSC).
The contributions of both funding sources were indispensable to the completion of this research.
\end{acks}

\bibliographystyle{ACM-Reference-Format}
\bibliography{sample-base}

@String{Computing = "Computing" }

@String{Springer = "Springer-Verlag" }

@ArtifactSoftware{R,
    title = {R: A Language and Environment for Statistical Computing},
    author = {{R Core Team}},
    organization = {R Foundation for Statistical Computing},
    address = {Vienna, Austria},
    year = {2019},
    url = {https://www.R-project.org/},
}

@article{DAMR,
  title={DAMR: Dynamic Adjacency Matrix Representation Learning for Multivariate Time Series Imputation},
  author={Ren, Xiaobin and Zhao, Kaiqi and Riddle, Patricia J and Taskova, Katerina and Pan, Qingyi and Li, Lianyan},
  journal={Proceedings of the ACM on Management of Data},
  volume={1},
  number={2},
  pages={1--25},
  year={2023},
  publisher={ACM New York, NY, USA}
}

@inproceedings{INCREASE,
title={Increase: Inductive graph representation learning for spatio-temporal kriging},
author={Zheng, Chuanpan and Fan, Xiaoliang and Wang, Cheng and Qi, Jianzhong and Chen, Chaochao and Chen, Longbiao},
booktitle={Proceedings of the ACM Web Conference 2023},
pages={673--683},
year={2023}
}

@article{KrigingReview,
  title={Spatial interpolation methods applied in the environmental sciences: A review},
  author={Li, Jin and Heap, Andrew D},
  journal={Environmental Modelling \& Software},
  volume={53},
  pages={173--189},
  year={2014},
  publisher={Elsevier}
}

@article{kriging1990,
  title={Kriging: a method of interpolation for geographical information systems},
  author={Oliver, Margaret A and Webster, Richard},
  journal={International Journal of Geographical Information System},
  volume={4},
  number={3},
  pages={313--332},
  year={1990},
  publisher={Taylor \& Francis}
}

@article{adaptiveIDW,
  title={An adaptive inverse-distance weighting spatial interpolation technique},
  author={Lu, George Y and Wong, David W},
  journal={Computers \& geosciences},
  volume={34},
  number={9},
  pages={1044--1055},
  year={2008},
  publisher={Elsevier}
}

@inproceedings{KCN,
  title={Kriging convolutional networks},
  author={Appleby, Gabriel and Liu, Linfeng and Liu, Li-Ping},
  booktitle={Proceedings of the AAAI Conference on Artificial Intelligence},
  volume={34},
  number={04},
  pages={3187--3194},
  year={2020}
}

@article{SSIN,
  title={SSIN: Self-Supervised Learning for Rainfall Spatial Interpolation},
  author={Li, Jia and Shen, Yanyan and Chen, Lei and Ng, Charles Wang Wai},
  journal={Proceedings of the ACM on Management of Data},
  volume={1},
  number={2},
  pages={1--21},
  year={2023},
  publisher={ACM New York, NY, USA}
}

@inproceedings{IGNNK,
  title={Inductive graph neural networks for spatiotemporal kriging},
  author={Wu, Yuankai and Zhuang, Dingyi and Labbe, Aurelie and Sun, Lijun},
  booktitle={Proceedings of the AAAI Conference on Artificial Intelligence},
  volume={35},
  number={5},
  pages={4478--4485},
  year={2021}
}

@article{GAT,
  title={Graph attention networks},
  author={Veli{\v{c}}kovi{\'c}, Petar and Cucurull, Guillem and Casanova, Arantxa and Romero, Adriana and Lio, Pietro and Bengio, Yoshua},
  journal={International Conference on Learning Representations},
  year={2018}
}

@inproceedings{graphsage,
author = {Hamilton, William L. and Ying, Rex and Leskovec, Jure},
title = {Inductive representation learning on large graphs},
year = {2017},
address = {Red Hook, NY, USA},
booktitle = {Proceedings of the 31st International Conference on Neural Information Processing Systems},
pages = {1025–1035},
numpages = {11},
location = {Long Beach, California, USA},
series = {NIPS'17}
}

@article{grin,
  title={Filling the G\_AP\_S: Multivariate time series imputation by graph neural networks},
  author={Cini, Andrea and Marisca, Ivan and Alippi, Cesare},
  journal={International Conference on Learning Representations},
  year={2022},
}

@article{spabert,
  title={SpaBERT: A pretrained language model from geographic data for geo-entity representation},
  author={Li, Zekun and Kim, Jina and Chiang, Yao-Yi and Chen, Muhao},
  journal={EMNLP},
  year={2022}
}

@article{largeconvex,
  title={Finding the convex hull of a simple polygon},
  author={Graham, Ronald L and Yao, F Frances},
  journal={Journal of Algorithms},
  volume={4},
  number={4},
  pages={324--331},
  year={1983},
  publisher={Elsevier}
}

@article{pointtheory,
  title={Point pattern analysis},
  author={Boots, Barry N and Getls, Arthur},
  year={2020},
  publisher={Regional Research Institute, West Virginia University}
}

@article{soil_moisture,
  title={Comparison of spatial interpolation methods for soil moisture and its application for monitoring drought},
  author={Chen, Hui and Fan, Li and Wu, Wei and Liu, Hong-Bin},
  journal={Environmental monitoring and assessment},
  volume={189},
  pages={1--13},
  year={2017},
  publisher={Springer}
}

@article{ok,
  title={Ordinary kriging},
  author={Wackernagel, Hans and Wackernagel, Hans},
  journal={Multivariate geostatistics: an introduction with applications},
  pages={79--88},
  year={2003},
  publisher={Springer}
}

@article{Herbold2020,
author = {Herbold, Steffen},
journal = {Journal of Open Source Software},
number = {48},
pages = {2173},
title = {{Autorank: A Python package for automated ranking of classifiers}},
volume = {5},
year = {2020}
}

@article{MCMC,
  title={An introduction to MCMC for machine learning},
  author={Andrieu, Christophe and De Freitas, Nando and Doucet, Arnaud and Jordan, Michael I},
  journal={Machine learning},
  volume={50},
  pages={5--43},
  year={2003},
  publisher={Springer}
}

@inproceedings{SZ_source,
  title={Forecasting fine-grained air quality based on big data},
  author={Zheng, Yu and Yi, Xiuwen and Li, Ming and Li, Ruiyuan and Shan, Zhangqing and Chang, Eric and Li, Tianrui},
  booktitle={Proceedings of the 21th ACM SIGKDD international conference on knowledge discovery and data mining},
  pages={2267--2276},
  year={2015}
}

@inproceedings{UKF,
  title={The unscented Kalman filter for nonlinear estimation},
  author={Wan, Eric A and Van Der Merwe, Rudolph},
  booktitle={Proceedings of the IEEE 2000 adaptive systems for signal processing, communications, and control symposium},
  pages={153--158},
  year={2000},
  organization={IEEE}
}

@inproceedings{DGNN,
  title={Dynamic and multi-faceted spatio-temporal deep learning for traffic speed forecasting},
  author={Han, Liangzhe and Du, Bowen and Sun, Leilei and Fu, Yanjie and Lv, Yisheng and Xiong, Hui},
  booktitle={Proceedings of the 27th ACM SIGKDD conference on knowledge discovery \& data mining},
  pages={547--555},
  year={2021}
}

@article{MOE,
  title={Outrageously large neural networks: The sparsely-gated mixture-of-experts layer},
  author={Shazeer, Noam and Mirhoseini, Azalia and Maziarz, Krzysztof and Davis, Andy and Le, Quoc and Hinton, Geoffrey and Dean, Jeff},
  journal={arXiv preprint arXiv:1701.06538},
  year={2017}
}

@article{ADJdeterminant,
  title={The determinant of the adjacency matrix of a graph},
  author={Harary, Frank},
  journal={Siam Review},
  volume={4},
  number={3},
  pages={202--210},
  year={1962},
  publisher={SIAM}
}

@inproceedings{STGNP,
  title={Graph Neural Processes for Spatio-Temporal Extrapolation},
  author={Hu, Junfeng and Liang, Yuxuan and Fan, Zhencheng and Chen, Hongyang and Zheng, Yu and Zimmermann, Roger},
  booktitle={Proceedings of the 29th ACM SIGKDD Conference on Knowledge Discovery and Data Mining},
  pages={752--763},
  year={2023}
}

@article{GLTL,
  title={Fast multivariate spatio-temporal analysis via low rank tensor learning},
  author={Bahadori, Mohammad Taha and Yu, Qi Rose and Liu, Yan},
  journal={Advances in neural information processing systems},
  volume={27},
  year={2014}
}

@inproceedings{KITS,
  title={Kits: Inductive spatio-temporal kriging with increment training strategy},
  author={Xu, Qianxiong and Long, Cheng and Li, Ziyue and Ruan, Sijie and Zhao, Rui and Li, Zhishuai},
  booktitle={Proceedings of the AAAI Conference on Artificial Intelligence},
  volume={39},
  number={12},
  pages={12945--12953},
  year={2025}
}

@inproceedings{STFNN,
author = {Feng, Yutong and Wang, Qiongyan and Xia, Yutong and Huang, Junlin and Zhong, Siru and Liang, Yuxuan},
title = {Spatio-temporal field neural networks for air quality inference},
year = {2024},
booktitle = {Proceedings of the Thirty-Third International Joint Conference on Artificial Intelligence},
numpages = {9},
location = {Jeju, Korea},
series = {IJCAI '24}
}

@article{GHM,
  title={A generalized heterogeneity model for spatial interpolation},
  author={Luo, Peng and Song, Yongze and Zhu, Di and Cheng, Junyi and Meng, Liqiu},
  journal={International Journal of Geographical Information Science},
  volume={37},
  number={3},
  pages={634--659},
  year={2023},
  publisher={Taylor \& Francis}
}

@incollection{SpaAuto,
  title={Spatial autocorrelation},
  author={Getis, Arthur},
  booktitle={Handbook of applied spatial analysis: Software tools, methods and applications},
  pages={255--278},
  year={2009},
  publisher={Springer}
}

@article{STK,
  title={Geographical detector-based stratified regression kriging strategy for mapping soil organic carbon with high spatial heterogeneity},
  author={Liu, Yaolin and Chen, Yiyun and Wu, Zihao and Wang, Bozhi and Wang, Shaochen},
  journal={Catena},
  volume={196},
  pages={104953},
  year={2021},
  publisher={Elsevier}
}

@article{KED,
  title={Spatial interpolation of monthly climate data for Finland: comparing the performance of kriging and generalized additive models},
  author={Aalto, Juha and Pirinen, Pentti and Heikkinen, Juha and Ven{\"a}l{\"a}inen, Ari},
  journal={Theoretical and applied climatology},
  volume={112},
  pages={99--111},
  year={2013},
  publisher={Springer}
}

@article{PMSN,
  title={A simple method for principal strata effects when the outcome has been truncated due to death},
  author={Chiba, Yasutaka and VanderWeele, Tyler J},
  journal={American journal of epidemiology},
  volume={173},
  number={7},
  pages={745--751},
  year={2011},
  publisher={Oxford University Press}
}

@article{gridkriging,
  title={Geo-spatial grid-based transformations of precipitation estimates using spatial interpolation methods},
  author={Teegavarapu, Ramesh SV and Meskele, Tadesse and Pathak, Chandra S},
  journal={Computers \& Geosciences},
  volume={40},
  pages={28--39},
  year={2012},
  publisher={Elsevier}
}

@article{voronoi,
  title={Voronoi Diagrams.},
  author={Aurenhammer, Franz and Klein, Rolf},
  journal={Handbook of computational geometry},
  volume={5},
  number={10},
  pages={201--290},
  year={2000}
}

@article{SSH,
  title={A measure of spatial stratified heterogeneity},
  author={Wang, Jin-Feng and Zhang, Tong-Lin and Fu, Bo-Jie},
  journal={Ecological indicators},
  volume={67},
  pages={250--256},
  year={2016},
  publisher={Elsevier}
}

@article{satcn,
  title={Spatial aggregation and temporal convolution networks for real-time kriging},
  author={Wu, Yuankai and Zhuang, Dingyi and Lei, Mengying and Labbe, Aurelie and Sun, Lijun},
  journal={arXiv preprint arXiv:2109.12144},
  year={2021}
}

@ARTICLE{Imbalancefeatures,
  author={Jamshed, Muhammad Ali and Ali, Kamran and Abbasi, Qammer H. and Imran, Muhammad Ali and Ur-Rehman, Masood},
  journal={IEEE Sensors Journal}, 
  title={Challenges, Applications, and Future of Wireless Sensors in Internet of Things: A Review}, 
  year={2022},
  volume={22},
  number={6},
  pages={5482-5494},
}

@inproceedings{NDBC,
  title={Automated data quality assurance for marine observations},
  author={Koziana, James V and Olson, John and Anselmo, Troy and Lu, William},
  booktitle={OCEANS 2008},
  pages={1--6},
  year={2008},
  organization={IEEE}
}

@inproceedings{NDBC2,
  title={Evaluating FY-3E GNOS-II Global Wind Product for Nearshore and Open Ocean: A Study Utilizing NDBC and TAO/TRITON Buoy Data},
  author={Han, Xinhai and Li, Xiaohui and Yang, Jingsong and Tao, Wei and Wang, Yiqi},
  booktitle={IGARSS 2024-2024 IEEE International Geoscience and Remote Sensing Symposium},
  pages={6372--6375},
  year={2024},
  organization={IEEE}
}

@article{Gridsearch,
  title={Grid search, random search, genetic algorithm: a big comparison for NAS},
  author={Liashchynskyi, Petro and Liashchynskyi, Pavlo},
  journal={arXiv preprint arXiv:1912.06059},
  year={2019}
}

@article{Akinshin,
    author = {Kruschke, John K. and Liddell, Torrin M.},
    title = {The Bayesian New Statistics: Hypothesis testing, estimation, meta-analysis, and power analysis from a Bayesian perspective},
    journal = {Psychonomic Bulletin \& Review},
    volume = {25},
    number = {1},
    pages = {178--206},
    year = {2018},
    month = {Feb},
    issn = {1531-5320},
}

\appendix

\section{Table of variables}\label{sec:Tab_of_variables}

\begin{table}[h]
    \centering
    \caption{Summary of Symbols}
    \begin{tabular}{cl}
        \hline
        \textbf{Symbol} & \textbf{Description} \\
        \hline
        $A$ & Adjacency matrix in Graph Neural Networks \\
        $a$ & Element in the adjacency matrix \\
        $D$ & Distance matrix for constructing adjacency matrix \\
        $d_{m,n}$ & Distance between unobserved and observed locations \\
        $E$ & Number of experts \\
        $F$ & Number of features \\
        $f$ & $f$-th feature index \\
        $G_p$ & $p$-th gate component \\
        $H$ & Latent representation \\
        $i, j$ & $i$-th and $j$-th location index \\
        $K$ & Set of neighboring locations selected in Eq.1 \\
        $k$ & $k$-th anchor location index \\
        $L$ & Location set (latitude and longitude) \\
        $M$ & Number of unobserved locations \\
        $N$ & Number of observed locations \\
        $P$ & Set of grid cells \\
        $Q$ & Number of anchor locations \\
        $r_{obs}$ & Observed average nearest neighbor distance \\
        $r_{exp}$ & Expected nearest neighbor distance \\
        $S$ & location set selected in strata creation \\
        $T$ & Length of time window \\
        $U$ & Number of neighboring locations \\
        $W_c^{(i)}$ & UKF covariance weights for sigma points \\
        $W_m^{(i)}$ & UKF mean weights for sigma points \\
        $\alpha, \beta, \kappa$ & UKF parameters for sigma point distribution \\
        $\gamma$ & Sigma points in the measurement space \\
        $w$ & Learnable weights \\
        $\rho$ & Pearson correlation coefficient \\
        $\epsilon$ & Threshold for sparsity \\
        $\mathcal{X}$ & Input time series \\
        $\mathcal{Y}$ & Output time series \\
        $\omega$ & Area in the strata \\
        $\sigma$ & Threshold for distribution \\
        \hline
    \end{tabular}
    \label{tab:symbols}
\end{table}

\section{Additional details on SSCC}\label{sec:TopK_Pearson}

Next, we show that Top-\(K\) Pearson correlation coefficient achieves the highest intra-strata homogeneity. Let $S_{k}$ be the $K$ locations selected by Eq.~\ref{eq:location_list} and $r_{ij}$ be the Pearson correlation coefficient between the $i$-th and the $j$-th locations, the intra-strata homogeneity $H_\text{intra}(S_{k})$ is:

\begin{equation}
H_\text{intra}(S_k) = \frac{1}{K(K-1)} \sum_{i \neq j, i,j \in S_k} r_{ij}.
\end{equation}

The Top-\(K\) Pearson correlation coefficient selection ensures the locations in \( S_k \) have the highest correlation values with the anchor location \( x_k \), i.e., \( r_{k,i} \geq r_{k,j} \) for all \( x_{k,i} \in S_k \) and \( x_{k,j} \notin S_k \). Since \( r_{ij} \leq \min(r_{k,i}, r_{k,j}) \) by the properties of the Pearson correlation coefficient, selecting \( K \) most correlated locations maximizes the pairwise correlations in \( S_k \), ensuring \( H_\text{intra}(S_k) \) is maximized. Adding any location \( x_{j} \notin S_k \) would reduce the mean correlation due to its lower similarity with the anchor location, violating the optimality of the selection.

\section{Details of Cross-Feature Estimator algorithm}\label{alg:CFE}

Algorithm~\label{alg:CFE} summarizes the procedure of the proposed Cross-Feature Estimator. Lines 1–7 initialize the learnable modules (FFN1, FFN2, and fusion weights), the UKF parameters, and the Merwe-scaled sigma-point weights. Lines 9–11 describe the one-step ahead prediction, where the UKF performs the prediction step and the two feature sources are transformed and fused into a unified measurement. Lines 12–16 correspond to the UKF update stage, including the computation of the predicted measurement, measurement covariance, cross-covariance, Kalman gain, and the subsequent update of the state mean and covariance. Finally, Line 17 stores the estimated state at each time step. This process enables the model to adaptively combine local and global feature information while maintaining uncertainty-aware state estimation.

\begin{algorithm}
\caption{The algorithm of Cross-Feature Estimator}
\label{alg:ukf_weighted_sum}

\begin{algorithmic}[1]
\STATE \textbf{Input:} \( H_{k}^{m} \in \mathbb{R}^{F' \times T \times j} \), \( \overline{X}^{global} \in \mathbb{R}^{T \times j} \)
\STATE \textbf{Output:} Estimated states \(\tilde{H}_{k}^{m} \in \mathbb{R}^{T \times j}\)

\STATE Initialize FFN models \( \text{FFN1} \) and \( \text{FFN2} \), learnable weights \( w \in \mathbb{R}^{T \times j} \), UKF weights  \(W_{m}^{(i)}\) and \(W_{c}^{(i)} \),  state \( x_0 = \mathbf{0} \in \mathbb{R}^{j} \), covariance \( P_0 = I_j \in \mathbb{R}^{j \times j} \)                                                                            
\STATE Set UKF parameters \( \alpha = 0.1 \), \( \beta = 2.0 \), \( \kappa = 0 \), and Merwe scaled sigma points for state dimension \( j \)
\STATE Compute \( \xi = \alpha^2 (j + \kappa) - j \)
\STATE Compute weights \( W_m^{(0)} = \frac{\xi}{j + \xi} \), \( W_c^{(0)} = \frac{\xi}{j + \xi} + (1 - \alpha^2 + \beta) \)
\STATE For \( i = 1, \ldots, 2j \): set \( W_m^{(i)} = W_c^{(i)} = \frac{1}{2(j + \xi)} \)

\FOR{$t = 0$ to $T-1$}
    \STATE Predict step: \( \text{UKF.predict()} \)
    \STATE Transform inputs: \( h_k = \text{FFN1}(H_{k}^{m}[:, t, :].\text{reshape}(-1)) \), \( x_k = \text{FFN2}(\overline{X}^{global}[t, :]) \)
    \STATE Compute weighted sum: \( z_k = w_t \odot h_k + (1 - w_t) \odot x_k \)
    \STATE Update step with the fused measurement:
    \STATE \textbf{Measurement Prediction:} 
    \[
    \hat{z}_k = \sum_{i=0}^{2L} W_m^{(i)} \gamma_k^{(i)}, \quad P_{zz} = \sum_{i=0}^{2L} W_c^{(i)} (\gamma_k^{(i)} - \hat{z}_k)(\gamma_k^{(i)} - \hat{z}_k)^T + R_k
    \]
    \STATE \textbf{Cross Covariance:} 
    \[
    P_{xz} = \sum_{i=0}^{2L} W_c^{(i)} (\chi_k^{(i)} - \hat{x}_k)(\gamma_k^{(i)} - \hat{z}_k)^T
    \]
    \STATE \textbf{Kalman Gain:} 
    \[
    K_k = P_{xz} P_{zz}^{-1}
    \]
    \STATE \textbf{State and Covariance Update:} 
    \[
    \hat{x}_k = \hat{x}_k + K_k (z_k - \hat{z}_k), \quad P_k = P_k - K_k P_{zz} K_k^T
    \]
    \STATE Store the estimated state: \( \tilde{H}_{k}^{m}[t, :] = \text{UKF.x} \)
\ENDFOR

\STATE \textbf{Return:} Estimated states \(\tilde{H}_{k}^{m}\)

\end{algorithmic}
\end{algorithm}

\section{Training and inference workflow}

\label{sec:Train_inference_example}

\begin{figure*}
    \centering
    \includegraphics[width=\linewidth]{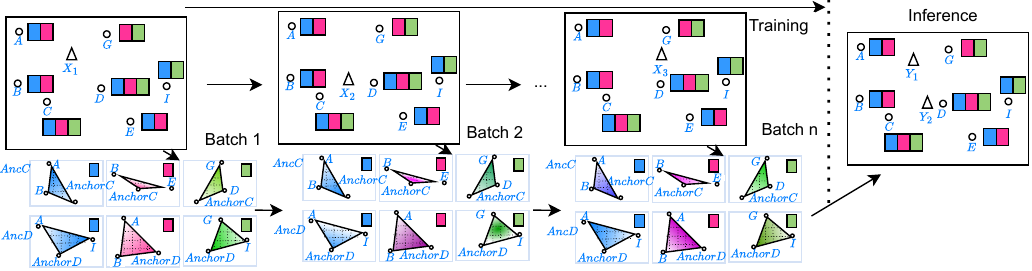}
    \caption{Illustration of the procedure of updating spatial correlations on training graphs through an inductive way. At each training batch, a random subset of locations is masked, and the model learns to reconstruct their values using spatial correlations across strata. During inference, the learned graph and inter-stratum dependencies enable estimation at previously unobserved locations.
}
    \label{fig:train_strategy}
\end{figure*}

To better illustrate our training and inference workflow, we provide Figure~\ref{fig:train_strategy}. At each training batch, we randomly mask a subset of spatial locations and require the model to reconstruct their values. This procedure induces continual updates of spatial correlations within and across strata, allowing the adjacency representation to evolve dynamically throughout training. As discussed in Section~\ref{sec:adj_rep}, the spatial correlation estimation is designed to be inductive: the model learns to leverage both intra-stratum and inter-stratum dependencies to propagate information from the observed locations to the unobserved locations. During inference, the learned spatial graph structure and the cross-stratum correlation patterns enable the model to generalize smoothly to previously unseen or even out-of-stratum locations, thereby supporting robust and flexible spatial prediction.

\end{document}